\documentclass[12pt,a4paper]{article}

\usepackage[T1]{fontenc}
\usepackage[margin=1in]{geometry}
\usepackage{amsmath,amssymb,amsfonts}
\usepackage{bm}
\usepackage{graphicx}
\usepackage{booktabs}
\usepackage{array}
\usepackage{multirow}
\usepackage{float}
\usepackage{caption}
\usepackage{subcaption}
\usepackage{siunitx}
\usepackage{cite}
\usepackage{xcolor}
\usepackage{hyperref}
\usepackage{enumitem}
\usepackage{authblk}
\usepackage{microtype}

\hypersetup{
	colorlinks=true,
	linkcolor=blue,
	citecolor=blue,
	urlcolor=blue
}

\allowdisplaybreaks

\title{
	\textbf{Adaptive Physics-Informed Neural Networks
		for the Blasius Boundary-Layer Problem}
}

\author[]{Mehari Fentahun Endalew}
\author[]{Xiaoming John Zhang}

\affil[]{
	Beijing Institute of Mathematical Sciences and Applications (BIMSA),
	Beijing, China
}

\date{}

\begin{document}
	
	\maketitle
	
	\begin{abstract}
		Physics-informed neural networks (PINNs) provide a mesh-free approach
		for solving differential equations, but their performance can depend
		strongly on loss weighting, collocation placement, and optimization
		strategy. This study develops an adaptive PINN framework for the Blasius
		boundary-layer equation using gradient-norm-based adaptive loss
		weighting, nonuniform and residual-based collocation, and sequential
		Adam--L-BFGS optimization. In the representative run using the
		architecture $[1,100,100,1]$, the model predicts
		$f''(0)=0.3320762918$, compared with the high-accuracy benchmark
		$0.332057336215$, giving an absolute error of
		$1.896\times10^{-5}$. The final weighted loss is
		$6.789\times10^{-8}$, and the predicted stream-function, velocity, and
		shear profiles agree closely with an independent numerical
		boundary-value solution. A separate full-training architecture study
		shows that the two-hidden-layer model achieves the smallest wall-shear
		error among the four tested architectures, $1.629\times10^{-6}$,
		whereas the deepest network attains the smallest weighted objective
		but a substantially larger wall-shear error. Compared with the
		previously reported PINN value $f''(0)=0.33165$, the representative
		run reduces the wall-shear error by approximately a factor of $21.5$.
		The results show that the combined adaptive training framework can
		achieve high accuracy for the Blasius problem and that weighted loss
		alone is insufficient for identifying the most physically accurate
		PINN. Because the adaptive components are applied jointly, their
		individual contributions cannot be isolated from the present results
		and would require a controlled ablation study for separate assessment.
	\end{abstract}


\section{Introduction}
\label{sec:introduction}

The laminar boundary layer over a flat plate is a classical problem in
fluid mechanics. Blasius~\cite{blasius1908} introduced a similarity
transformation that reduces the steady, incompressible,
two-dimensional boundary-layer equations for zero pressure-gradient
flow to a third-order nonlinear ordinary differential equation. The
resulting Blasius solution provides a fundamental description of
laminar boundary-layer development and remains an important benchmark
for analytical, numerical, and computational methods. Howarth
~\cite{howarth1938} subsequently provided detailed numerical solutions
of the laminar boundary-layer equations and tabulated the corresponding
Blasius velocity profile.

An important quantity associated with the Blasius solution is the
wall-shear parameter $f''(0)$, which is directly related to the
skin-friction coefficient. In the present study, the high-accuracy
reference value

\begin{equation}
	f''(0)_{\mathrm{ref}}
	=
	0.332057336215
	\label{eq:benchmark}
\end{equation}

is used as the principal scalar quantity for evaluating the accuracy
of the computed solution~\cite{boyd1999,lalmartin2022}.

Traditional numerical treatments of the Blasius problem commonly use
shooting, finite-difference, finite-element, or other
discretization-based techniques. Physics-informed neural networks
(PINNs) provide an alternative approach in which the governing
differential equation and associated boundary conditions are embedded
directly into the neural-network optimization objective. Raissi
et al.~\cite{raissi2019} established the modern PINN framework for
solving forward and inverse problems governed by nonlinear differential
equations by combining neural-network approximation with automatic
differentiation and physics-based residual losses.

The development of scientific-machine-learning software has further
facilitated the application of PINNs to differential equations. For
example, Lu et al.~\cite{lu2021} developed DeepXDE, a general-purpose
deep-learning library for solving differential equations using
physics-informed and related neural-network approaches. Such
developments have contributed to the broader adoption of PINNs in
computational science. The present implementation, however, is
developed directly in PyTorch to provide explicit control over the
adaptive loss-weighting and collocation procedures.

Recent studies also illustrate the application of PINNs to fluid-flow
and transport problems beyond the Blasius equation. Anthony et
al.~\cite{anthony2025} applied a PINN to the two-dimensional steady
incompressible Navier--Stokes equations with an application to
Poiseuille flow. Mathur et al.~\cite{mathur2026} investigated a
physics-informed neural-network formulation for radiative
magnetohydrodynamic mixed convection in reactive nanofluid slip flow
over a moving vertical plate. These studies further demonstrate the
use of physics-informed learning approaches for nonlinear fluid-flow
problems.

Despite their flexibility, PINNs can be difficult to train because the
different terms in a composite physics-informed objective may generate
gradients with substantially different magnitudes. Wang et
al.~\cite{wang2021} analyzed such gradient-flow pathologies and showed
that imbalance among loss components can significantly affect PINN
training. In addition, the placement of collocation points can influence
solution accuracy and convergence. Adaptive sampling strategies have
therefore been investigated to concentrate training points in regions
where the governing equations are more difficult to satisfy
~\cite{subramanian2022,khanra2026}. More generally, adaptive localized
treatments have also been explored for challenging nonlinear
physics-informed problems~\cite{coutinho2022}.

For the Blasius problem specifically, Krishna et
al.~\cite{krishna2023} applied a PINN directly to the Blasius
boundary-value problem. Their formulation used automatic
differentiation to construct the governing-equation residual and
employed $100$ equidistant collocation points over the truncated domain

\begin{equation}
	0\leq\eta\leq8.
\end{equation}

For a neural network with two hidden layers and $100$ neurons per
hidden layer, Krishna et al.~\cite{krishna2023} reported

\begin{equation}
	f''(0)=0.33165,
\end{equation}

with a reported loss of

\begin{equation}
	1.67\times10^{-6}.
\end{equation}

Their study also investigated the influence of neural-network
architecture and extended the PINN solution toward the negative
$\eta$-axis, where rapid growth near the known Blasius singularity
around $\eta=-5.69$ was reproduced
~\cite{boyd1999,krishna2023}.

Motivated by these developments and by the known sensitivity of PINNs
to gradient imbalance and collocation-point placement
~\cite{wang2021,subramanian2022}, the present study develops an
adaptive PINN for the Blasius boundary-layer problem. The method
combines gradient-norm-based adaptive loss weighting, nonuniform
near-wall collocation, residual-based adaptive refinement, and
sequential Adam--L-BFGS optimization. The resulting solution is
evaluated using the wall-shear parameter $f''(0)$,
boundary-condition errors, complete stream-function, velocity, and
shear profiles, an independent numerical boundary-value solution, and
the pointwise governing-equation residual. A separate full-training
architecture study is also performed to investigate the relationship
between network depth, weighted optimization loss, and physical
wall-shear accuracy.

	\section{Mathematical Formulation}
	\label{sec:mathematical}
	
	\subsection{Boundary-Layer Equations}
	
	Consider steady, incompressible, two-dimensional laminar flow over a
	flat plate. Let $u(x,y)$ and $v(x,y)$ denote the streamwise and
	wall-normal velocity components, respectively. Under the
	boundary-layer approximation, the governing equations are
	
	\begin{equation}
		\frac{\partial u}{\partial x}
		+
		\frac{\partial v}{\partial y}
		=
		0,
		\label{eq:continuity}
	\end{equation}
	
	and
	
	\begin{equation}
		u\frac{\partial u}{\partial x}
		+
		v\frac{\partial u}{\partial y}
		=
		\nu
		\frac{\partial^2u}{\partial y^2},
		\label{eq:momentum}
	\end{equation}
	
	where $\nu$ denotes the kinematic viscosity.
	
	The no-slip and impermeability conditions are
	
	\begin{equation}
		u(x,0)=0,
		\qquad
		v(x,0)=0,
	\end{equation}
	
	while the far-field velocity condition is
	
	\begin{equation}
		u(x,y)\rightarrow U_{\infty},
		\qquad
		y\rightarrow\infty.
	\end{equation}
	
	\subsection{Blasius Similarity Transformation}
	
	Introducing the similarity coordinate
	
	\begin{equation}
		\eta
		=
		y\sqrt{\frac{U_{\infty}}{\nu x}},
	\end{equation}
	
	and the stream function
	
	\begin{equation}
		\psi
		=
		\sqrt{\nu U_{\infty}x}\,
		f(\eta),
	\end{equation}
	
	reduces the boundary-layer equations to the classical Blasius equation
	
	\begin{equation}
		f'''(\eta)
		+
		\frac{1}{2}
		f(\eta)f''(\eta)
		=
		0.
		\label{eq:blasius}
	\end{equation}
	
	The corresponding boundary conditions are
	
	\begin{equation}
		f(0)=0,
		\qquad
		f'(0)=0,
		\qquad
		f'(\infty)=1.
		\label{eq:blasius_bc}
	\end{equation}
	
	The normalized streamwise velocity is
	
	\begin{equation}
		\frac{u}{U_{\infty}}
		=
		f'(\eta),
		\label{eq:velocity}
	\end{equation}
	
	while
	
	\begin{equation}
		\frac{\partial u}{\partial y}
		=
		U_{\infty}
		\sqrt{\frac{U_{\infty}}{\nu x}}
		f''(\eta).
		\label{eq:shear}
	\end{equation}
	
	Thus, $f''(0)$ determines the nondimensional wall-shear parameter and
	is used as the principal physical accuracy measure throughout this
	study.
	
\section{Physics-Informed Neural Network Formulation}
\label{sec:pinn}

Following the general PINN framework introduced by Raissi et
al.~\cite{raissi2019}, the unknown solution is represented by a neural
network, while the governing differential equation and boundary
conditions are incorporated into the training objective through
automatic differentiation and residual minimization. General-purpose
frameworks such as DeepXDE~\cite{lu2021} provide related tools for
physics-informed differential-equation solvers. In the present study,
the PINN is implemented directly in PyTorch to provide explicit control
over the adaptive loss-weighting and collocation procedures.

\subsection{Neural Approximation}

The Blasius similarity function is approximated by a fully connected
neural network,

\begin{equation}
	f(\eta)
	\approx
	f_{\theta}(\eta)
	=
	\mathrm{NN}_{\theta}(\eta),
	\label{eq:nn}
\end{equation}

where $\theta$ denotes all trainable weights and biases.

The representative network architecture is

\begin{equation}
	[1,100,100,1],
	\label{eq:architecture}
\end{equation}

with hyperbolic-tangent activation functions in the hidden layers.
Network weights are initialized using Xavier normal initialization,
while all biases are initialized to zero.

Automatic differentiation is used to evaluate

\begin{equation}
	f'_{\theta}(\eta),
	\qquad
	f''_{\theta}(\eta),
	\qquad
	f'''_{\theta}(\eta).
\end{equation}

Using these derivatives, the Blasius differential-equation residual is
defined as

\begin{equation}
	R_{\theta}(\eta)
	=
	f'''_{\theta}(\eta)
	+
	\frac{1}{2}
	f_{\theta}(\eta)
	f''_{\theta}(\eta).
	\label{eq:residual}
\end{equation}

\subsection{Computational Domain}

The semi-infinite Blasius domain is truncated to the finite interval

\begin{equation}
	\eta\in[0,8].
	\label{eq:domain}
\end{equation}

Accordingly, the asymptotic boundary condition
$f'(\infty)=1$ is approximated numerically by

\begin{equation}
	f'_{\theta}(8)=1.
	\label{eq:far_condition}
\end{equation}

\subsection{Loss Function}

The network parameters are determined by minimizing a weighted
physics-informed objective consisting of the governing-equation,
wall-boundary, and far-field contributions:

\begin{equation}
	\mathcal{L}
	=
	\lambda_{\mathrm{ODE}}
	\mathcal{L}_{\mathrm{ODE}}
	+
	\lambda_{\mathrm{Wall}}
	\mathcal{L}_{\mathrm{Wall}}
	+
	\lambda_{\mathrm{Far}}
	\mathcal{L}_{\mathrm{Far}}.
	\label{eq:total_loss}
\end{equation}

The ODE residual loss is

\begin{equation}
	\mathcal{L}_{\mathrm{ODE}}
	=
	\frac{1}{N_c}
	\sum_{i=1}^{N_c}
	\left[
	R_{\theta}(\eta_i)
	\right]^2,
	\label{eq:ode_loss}
\end{equation}

where $N_c$ denotes the number of collocation points.

The wall-boundary loss enforces the conditions
$f(0)=0$ and $f'(0)=0$:

\begin{equation}
	\mathcal{L}_{\mathrm{Wall}}
	=
	[f_{\theta}(0)]^2
	+
	[f'_{\theta}(0)]^2.
	\label{eq:wall_loss}
\end{equation}

The far-field contribution enforces the truncated condition
$f'_{\theta}(8)=1$:

\begin{equation}
	\mathcal{L}_{\mathrm{Far}}
	=
	[f'_{\theta}(8)-1]^2.
	\label{eq:far_loss}
\end{equation}

The coefficients
$\lambda_{\mathrm{ODE}}$,
$\lambda_{\mathrm{Wall}}$, and
$\lambda_{\mathrm{Far}}$
control the relative contributions of the three loss components and
are updated adaptively during Adam optimization, as described in
Section~\ref{sec:method}.
	\subsection{Relation to the Previous Blasius PINN}
	
	The present formulation retains the central PINN structure used by
	Krishna et al.~\cite{krishna2023}: direct neural approximation of the
	Blasius function, automatic differentiation, and enforcement of the
	governing equation and boundary conditions through the optimization
	objective.
	
	The principal differences concern the training procedure. Krishna
	et al.~\cite{krishna2023} used $100$ equidistant collocation points
	over $0\leq\eta\leq8$. The present method begins with $499$ unique
	nonuniform points and subsequently introduces locations associated
	with large residuals. In addition, the relative weights of the ODE,
	wall, and far-field terms are adapted during Adam training instead of
	remaining fixed throughout optimization.
	
\section{Adaptive PINN Method}
\label{sec:method}

The individual terms in a PINN objective can generate parameter
gradients with substantially different magnitudes, leading to imbalance
during optimization. Such gradient-flow difficulties have been
identified as an important challenge in PINN training
~\cite{wang2021}. To reduce this imbalance, the present method
dynamically adjusts the relative coefficients of the ODE, wall, and
far-field loss components according to their parameter-gradient
magnitudes.

\subsection{Gradient-Norm Adaptive Loss Weighting}

Let the trainable parameter tensors be

\begin{equation}
	\theta
	=
	\{
	\theta_1,\theta_2,\ldots,\theta_M
	\}.
\end{equation}

For each loss component $\mathcal{L}_k$, the implemented gradient scale
is defined as

\begin{equation}
	g_k
	=
	\sum_{\ell=1}^{M}
	\left\|
	\nabla_{\theta_{\ell}}
	\mathcal{L}_k
	\right\|_2,
	\qquad
	k\in
	\{
	\mathrm{ODE},
	\mathrm{Wall},
	\mathrm{Far}
	\}.
	\label{eq:gradient_norm}
\end{equation}

The provisional adaptive coefficients are then computed as

\begin{equation}
	\widetilde{\lambda}_{\mathrm{ODE}}
	=
	\frac{
		g_{\mathrm{Wall}}+g_{\mathrm{Far}}
	}{
		2g_{\mathrm{ODE}}
	},
	\label{eq:lambda_ode}
\end{equation}

\begin{equation}
	\widetilde{\lambda}_{\mathrm{Wall}}
	=
	\frac{
		g_{\mathrm{ODE}}+g_{\mathrm{Far}}
	}{
		2g_{\mathrm{Wall}}
	},
	\label{eq:lambda_wall}
\end{equation}

and

\begin{equation}
	\widetilde{\lambda}_{\mathrm{Far}}
	=
	\frac{
		g_{\mathrm{ODE}}+g_{\mathrm{Wall}}
	}{
		2g_{\mathrm{Far}}
	}.
	\label{eq:lambda_far}
\end{equation}

To avoid unstable updates when a gradient scale is extremely small, a
coefficient is updated only when its corresponding gradient scale
satisfies

\begin{equation}
	g_k>10^{-12}.
\end{equation}

The resulting coefficient is then clipped according to

\begin{equation}
	\lambda_k
	=
	\operatorname{clip}
	\left(
	\widetilde{\lambda}_k,
	0.1,
	10
	\right),
\end{equation}

so that

\begin{equation}
	0.1
	\leq
	\lambda_k
	\leq
	10.
	\label{eq:weight_bounds}
\end{equation}

The adaptive coefficients are updated every $100$ Adam iterations.
After completion of the Adam stage, the final adaptive coefficients are held fixed during the subsequent L-BFGS refinement.
	
	\subsection{Nonuniform Base Collocation}
	
	A nominal set of $500$ base collocation points is generated. To provide
	additional near-wall resolution, $60\%$ of the points are distributed
	uniformly over
	
	\begin{equation}
		0\leq\eta\leq3,
	\end{equation}
	
	while the remaining $40\%$ are distributed uniformly over
	
	\begin{equation}
		3\leq\eta\leq8.
	\end{equation}
	
	Thus,
	
	\begin{equation}
		N_{\mathrm{near}}=300,
		\qquad
		N_{\mathrm{far}}=200.
	\end{equation}
	
	Because $\eta=3$ occurs in both subsets and duplicate coordinates are
	removed, the actual initial collocation set contains
	
	\begin{equation}
		N_c=499
	\end{equation}
	
	unique points.
	

\subsection{Residual-Based Adaptive Collocation}

During Adam training, the current Blasius residual is evaluated at

\begin{equation}
	N_{\mathrm{cand}}=3000
\end{equation}

uniformly distributed candidate locations. The $200$ candidate points
with the largest absolute residual values are selected:

\begin{equation}
	\Omega_{\mathrm{adapt}}
	=
	\operatorname{TopK}_{200}
	\left\{
	|R_{\theta}(\eta)|:
	\eta\in\Omega_{\mathrm{cand}}
	\right\}.
	\label{eq:adaptive_points}
\end{equation}

At each adaptive update, the previously selected adaptive subset is
discarded and replaced by the newly identified high-residual points.
The collocation set is therefore reconstructed from the fixed
nonuniform base set and the current adaptive subset according to

\begin{equation}
	\Omega_c
	=
	\operatorname{unique}
	\left(
	\Omega_{\mathrm{base}}
	\cup
	\Omega_{\mathrm{adapt}}
	\right).
\end{equation}

Thus, the adaptive points are replaced rather than accumulated across
successive refinement steps. Adaptive collocation updates are performed
at Adam iterations

\begin{equation}
	3000,\qquad
	6000,\qquad
	9000,\qquad
	12000.
	\label{eq:update_iterations}
\end{equation}

For the representative computation, each such reconstruction produced
$697$ unique collocation points after duplicate coordinates were
removed.

Before the L-BFGS stage, one additional residual-based refinement is
performed. The residual is evaluated at $4000$ uniformly distributed
candidate locations, and the $250$ points with the largest absolute
residual values are selected. The final collocation set used for
L-BFGS refinement is then reconstructed by combining these selected
points with the fixed nonuniform base set and removing duplicate
coordinates.
```

	\subsection{Adam Optimization}
	
	Adam optimization is performed for
	
	\begin{equation}
		N_{\mathrm{Adam}}=15000
	\end{equation}
	
	iterations with an initial learning rate
	
	\begin{equation}
		\alpha_{\mathrm{Adam}}=10^{-3}.
	\end{equation}
	
	A \texttt{ReduceLROnPlateau} scheduler is applied with patience $500$
	and reduction factor $0.5$. The norm of the full parameter gradient is
	clipped to a maximum value of $1$ before each Adam update.
	
	The implementation stores the network parameters whenever a new
	minimum weighted Adam loss is obtained. The minimum stored value in
	the representative run is
	
	\begin{equation}
		\mathcal{L}_{\mathrm{Adam,best}}
		=
		1.43\times10^{-9}.
		\label{eq:best_adam}
	\end{equation}
	
	The associated network parameters are restored before the final
	residual-based collocation refinement and L-BFGS optimization.
	
	Because both adaptive coefficients and collocation locations evolve
	during Adam training, weighted loss values from different stages do
	not necessarily represent identical effective optimization
	objectives. The stored minimum is therefore treated primarily as a
	training diagnostic.
	
	\subsection{L-BFGS Refinement}
	
	The restored Adam network is further refined using the PyTorch L-BFGS
	optimizer with learning rate
	
	\begin{equation}
		\alpha_{\mathrm{LBFGS}}=0.5.
	\end{equation}
	
	The optimizer settings are
	
	\begin{equation}
		\texttt{max\_iter}=100,
		\qquad
		\texttt{max\_eval}=200,
	\end{equation}
	
	with history size $50$, strong-Wolfe line search, gradient tolerance
	$10^{-12}$, and parameter-change tolerance $10^{-16}$.
	
	L-BFGS is invoked within an outer loop of at most $100$ calls. The
	outer process is terminated when
	
	\begin{equation}
		\left|
		\mathcal{L}^{(m)}
		-
		\mathcal{L}^{(m-1)}
		\right|
		<
		10^{-14},
		\label{eq:lbfgs_stop}
	\end{equation}
	
	or when the maximum number of outer calls is reached. The adaptive
	loss coefficients are not updated during this stage.
	
	\subsection{Computational Settings}
	
	The reported computation was performed using Python 3.12.8 on a
	64-bit Windows system with CPU execution. Double precision,
	
	\begin{equation}
		\texttt{torch.float64},
	\end{equation}
	
	was used throughout. The random-number generators were initialized
	using
	
	\begin{equation}
		\texttt{torch.manual\_seed}(42),
		\qquad
		\texttt{numpy.random.seed}(42).
	\end{equation}
	
	\section{Numerical Validation}
	\label{sec:validation}
	
\subsection{High-Accuracy Blasius Benchmark}

The principal scalar validation quantity is the high-accuracy
wall-shear value given in Eq.~\eqref{eq:benchmark},
consistent with high-precision results reported in the
literature~\cite{boyd1999,lalmartin2022}.
	
	\subsection{Independent Boundary-Value Solution}
	
	For complete profile validation, an independent finite-domain
	boundary-value solution is computed using SciPy's
	\texttt{solve\_bvp}. Introducing
	
	\begin{equation}
		y_1=f,
		\qquad
		y_2=f',
		\qquad
		y_3=f'',
	\end{equation}
	
	the first-order system becomes
	
	\begin{equation}
		y_1'=y_2,
		\qquad
		y_2'=y_3,
	\end{equation}
	
	and
	
	\begin{equation}
		y_3'
		=
		-\frac{1}{2}y_1y_3.
	\end{equation}
	
	The finite-domain boundary conditions are
	
	\begin{equation}
		y_1(0)=0,
		\qquad
		y_2(0)=0,
		\qquad
		y_2(8)=1.
	\end{equation}
	
	The numerical solver uses an initial mesh of $1000$ points, tolerance
	
	\begin{equation}
		10^{-10},
	\end{equation}
	
	and a maximum of $10000$ mesh nodes.
	
	The BVP solution is used to assess the complete PINN profiles, whereas
	the high-accuracy value in Eq.~\eqref{eq:benchmark} is used for the
	principal scalar wall-shear comparison.
	
	It should be noted that the high-accuracy reference corresponds to the
	semi-infinite Blasius problem, whereas both the PINN and the numerical
	BVP calculations impose the truncated condition $f'(8)=1$. Therefore,
	a small contribution to discrepancies relative to the semi-infinite
	reference may arise from finite-domain truncation in addition to
	neural-network approximation and optimization errors
	~\cite{fazio1992}.
	
	\section{Numerical Results}
	\label{sec:results}
	
	\subsection{Wall-Shear Accuracy}
	
	The representative adaptive PINN predicts
	
	\begin{equation}
		f''(0)_{\mathrm{PINN}}
		=
		0.3320762918.
		\label{eq:pinn_fpp}
	\end{equation}
	
	Relative to the reference value in Eq.~\eqref{eq:benchmark}, the
	absolute error is
	
	\begin{align}
		e_{\mathrm{abs}}
		&=
		\left|
		f''(0)_{\mathrm{PINN}}
		-
		f''(0)_{\mathrm{ref}}
		\right|
		\\
		&=
		1.896\times10^{-5}.
		\label{eq:absolute_error}
	\end{align}
	
	The corresponding relative error is approximately
	
	\begin{equation}
		e_{\mathrm{rel}}
		\approx
		0.00571\%.
	\end{equation}
	
	\subsection{Final Loss Components}
	
	The final loss components after L-BFGS refinement are summarized in
	Table~\ref{tab:final_loss}.
	
	\begin{table}[H]
		\centering
		\caption{Final loss values after L-BFGS refinement.}
		\label{tab:final_loss}
		\begin{tabular}{lc}
			\toprule
			\textbf{Quantity} & \textbf{Value} \\
			\midrule
			Total weighted loss & $6.789\times10^{-8}$ \\
			Raw ODE loss & $6.789\times10^{-9}$ \\
			Raw wall loss & $5.574\times10^{-12}$ \\
			Raw far-field loss & $1.435\times10^{-14}$ \\
			\bottomrule
		\end{tabular}
	\end{table}
	
	The wall and far-field losses are several orders of magnitude smaller
	than the raw ODE loss, indicating accurate enforcement of the
	prescribed boundary conditions.
	
	\subsection{Adam Training Behavior}
	
	Representative values from the Adam optimization stage are given in
	Table~\ref{tab:adam_history}. The adaptive loss coefficients are first
	evaluated at Adam iteration $0$, before the first parameter-update step,
	and are subsequently updated every $100$ Adam iterations, i.e., at iterations $0,100,200,\ldots$.
	
	\begin{table}[H]
		\centering
		\caption{Representative Adam training output.}
		\label{tab:adam_history}
		\resizebox{\textwidth}{!}{
			\begin{tabular}{cccccc}
				\toprule
				\textbf{Iteration} &
				\textbf{Weighted Loss} &
				$\bm{f''(0)}$ &
				$\bm{\lambda_{\mathrm{ODE}}}$ &
				$\bm{\lambda_{\mathrm{Wall}}}$ &
				$\bm{\lambda_{\mathrm{Far}}}$ \\
				\midrule
				0     & $9.61\times10^{-2}$ & 0.00130673 & 10.00 & 10.00 & 0.10 \\
				1000  & $1.29\times10^{-4}$ & 0.33104609 & 10.00 & 0.10 & 10.00 \\
				2000  & $1.74\times10^{-5}$ & 0.33163539 & 10.00 & 0.10 & 10.00 \\
				3000  & $1.13\times10^{-5}$ & 0.33170084 & 10.00 & 0.10 & 5.82 \\
				4000  & $7.53\times10^{-6}$ & 0.33247665 & 10.00 & 0.10 & 10.00 \\
				5000  & $3.21\times10^{-6}$ & 0.33231699 & 10.00 & 0.10 & 5.02 \\
				6000  & $1.24\times10^{-6}$ & 0.33221442 & 10.00 & 0.18 & 1.49 \\
				7000  & $7.75\times10^{-7}$ & 0.33197835 & 10.00 & 0.12 & 2.34 \\
				8000  & $4.87\times10^{-7}$ & 0.33204860 & 10.00 & 0.20 & 1.56 \\
				9000  & $1.74\times10^{-7}$ & 0.33204226 & 7.08 & 0.10 & 10.00 \\
				10000 & $1.99\times10^{-7}$ & 0.33210619 & 10.00 & 0.10 & 10.00 \\
				11000 & $1.45\times10^{-7}$ & 0.33210102 & 10.00 & 0.10 & 10.00 \\
				12000 & $1.14\times10^{-7}$ & 0.33209743 & 10.00 & 0.10 & 10.00 \\
				13000 & $5.31\times10^{-8}$ & 0.33205521 & 3.63 & 0.10 & 10.00 \\
				14000 & $1.43\times10^{-7}$ & 0.33205687 & 10.00 & 0.10 & 10.00 \\
				\bottomrule
			\end{tabular}
		}
	\end{table}
The wall-shear estimate generally approaches the reference value during
Adam optimization, with small fluctuations arising as training proceeds.
For example, the predictions at iterations $13000$ and $14000$ are

\begin{equation}
	f''(0)=0.33205521
	\qquad\text{and}\qquad
	f''(0)=0.33205687,
\end{equation}

respectively, both of which are close to the high-accuracy benchmark.
	
\begin{figure}[H]
		\centering
		\includegraphics[width=0.95\textwidth]{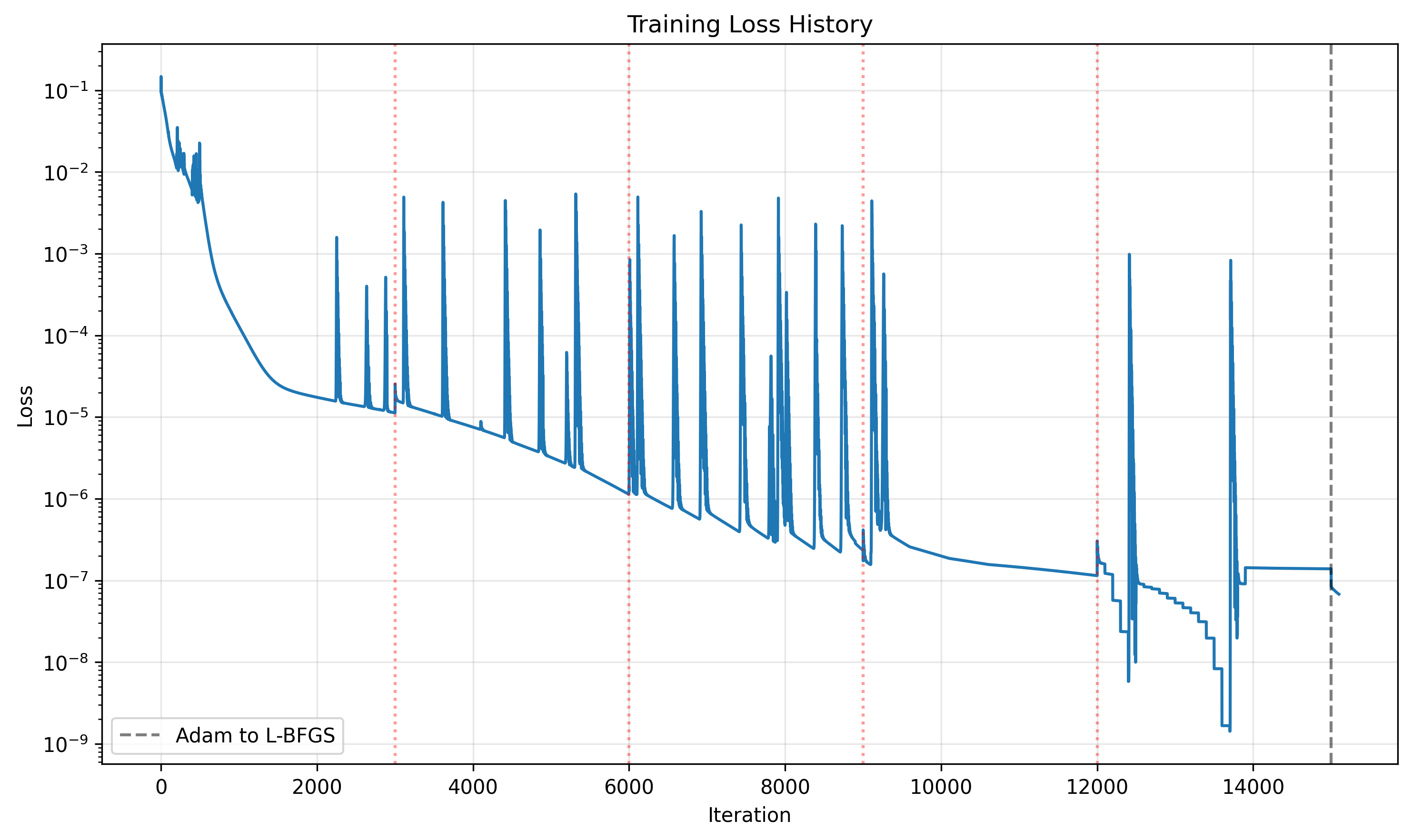}
		\caption{
			Training-loss history during Adam and L-BFGS optimization.
			The red dotted vertical lines denote residual-based collocation
			updates, while the black dashed line marks the transition from
			Adam to L-BFGS. The horizontal coordinate contains the Adam
			iterations followed by successive outer L-BFGS calls.
		}
		\label{fig:loss}
	\end{figure}

	The complete training trajectory is shown in
	Fig.~\ref{fig:loss}. The weighted objective decreases by several
	orders of magnitude, although short-lived upward changes occur. The
	red dotted vertical lines correspond to residual-based collocation
	updates at iterations $3000$, $6000$, $9000$, and $12000$.
	
	For an adaptive-update iteration, the tabulated loss is evaluated on
	the collocation set used for that Adam step before the subsequent
	adaptive set is constructed. Since newly introduced points are selected
	from regions with comparatively large residuals, the modified objective
	can become more demanding after refinement. A temporary increase in
	the weighted loss therefore does not necessarily imply deterioration of
	the physical solution.
	
	Additional changes in the loss arise from adaptive reweighting. Since
	the coefficients are updated every $100$ Adam iterations, the numerical
	value of the weighted objective may change when the relative
	contributions of the physical constraints are altered.
	
\begin{figure}[H]
		\centering
		\includegraphics[width=0.95\textwidth]{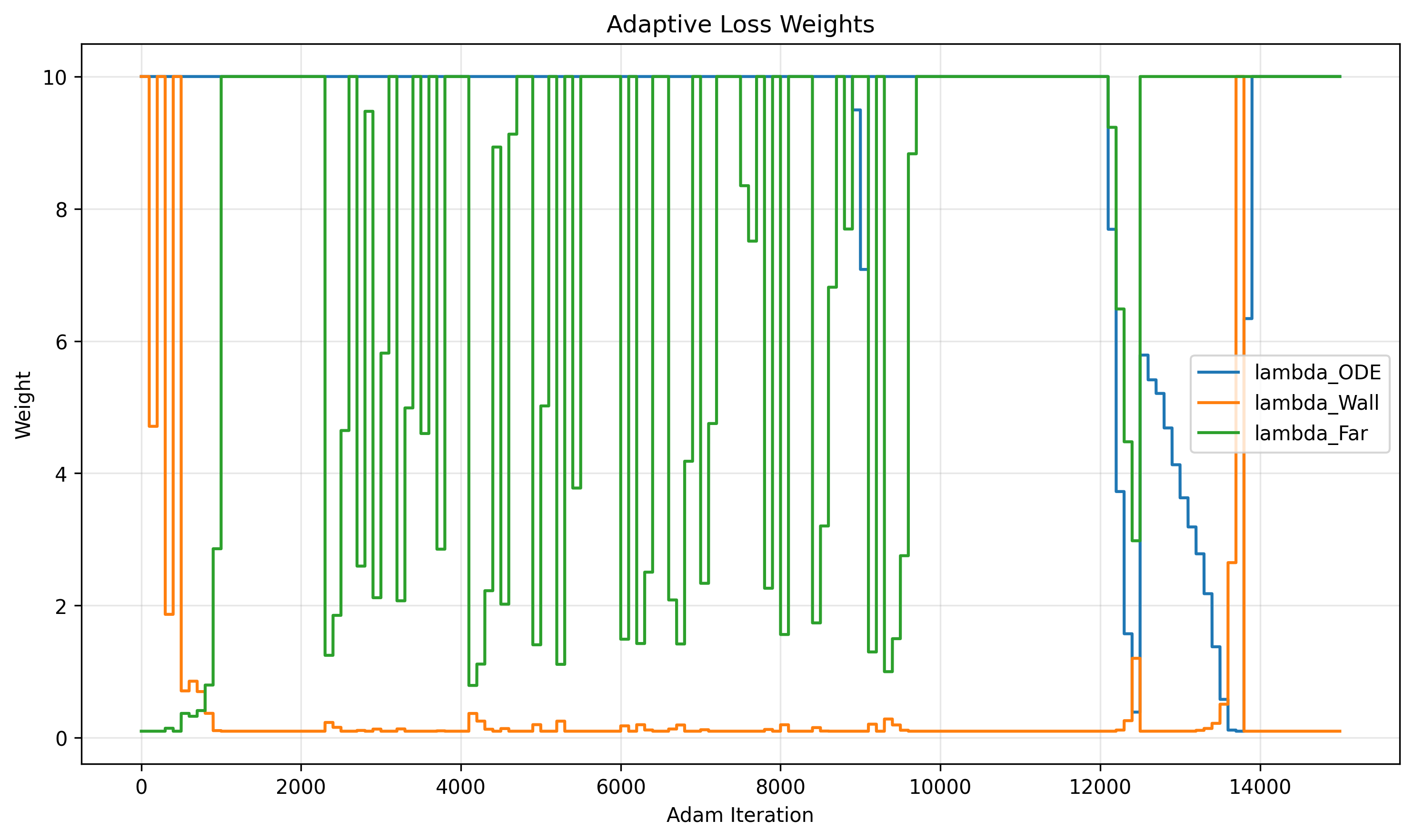}
		\caption{
			Evolution of the adaptive ODE, wall, and far-field loss
			coefficients during Adam training. The coefficients are restricted
			to $0.1\leq\lambda_k\leq10$.
		}
		\label{fig:weights}
	\end{figure}

	Figure~\ref{fig:weights} shows that the adaptive coefficients vary
	substantially during Adam optimization and frequently approach the
	imposed clipping limits. The wall coefficient is often close to the
	lower bound, whereas the ODE and far-field coefficients frequently
	take larger values.
	
	These coefficients should not be interpreted as direct measures of
	physical importance. Rather, they compensate for differences in the
	gradient scales generated by the individual loss components.
	
	\subsection{Transition from Adam to L-BFGS}
	
	The minimum weighted loss stored during Adam optimization is
	
	\begin{equation}
		1.43\times10^{-9}.
	\end{equation}
	
	The associated network parameters are restored before the final
	residual-based collocation refinement. After reconstruction of the
	collocation set, the first reported L-BFGS weighted loss is
	
	\begin{equation}
		8.50\times10^{-8}.
	\end{equation}
	
	These values should not be interpreted as evidence of deterioration,
	because they are evaluated under different collocation
	configurations. The adaptive coefficients also remain fixed during the
	L-BFGS stage. Thus, the second stage refines the restored model under
	the final training configuration rather than simply continuing an
	unchanged Adam objective.
	
	\subsection{L-BFGS Refinement}
	
	Representative L-BFGS results are shown in
	Table~\ref{tab:lbfgs_history}.
	
	\begin{table}[H]
		\centering
		\caption{Representative L-BFGS outer-step results.}
		\label{tab:lbfgs_history}
		\begin{tabular}{ccc}
			\toprule
			\textbf{Outer Step} &
			\textbf{Weighted Loss} &
			$\bm{f''(0)}$ \\
			\midrule
			0     & $8.50\times10^{-8}$ & 0.33207789 \\
			1     & $8.44\times10^{-8}$ & 0.33207733 \\
			2     & $8.39\times10^{-8}$ & 0.33207678 \\
			3     & $8.35\times10^{-8}$ & 0.33207634 \\
			4     & $8.31\times10^{-8}$ & 0.33207593 \\
			5     & $8.28\times10^{-8}$ & 0.33207566 \\
			6     & $8.25\times10^{-8}$ & 0.33207546 \\
			7     & $8.22\times10^{-8}$ & 0.33207531 \\
			8     & $8.20\times10^{-8}$ & 0.33207528 \\
			9     & $8.17\times10^{-8}$ & 0.33207523 \\
			10    & $8.15\times10^{-8}$ & 0.33207521 \\
			20    & $7.94\times10^{-8}$ & 0.33207539 \\
			30    & $7.75\times10^{-8}$ & 0.33207573 \\
			40    & $7.58\times10^{-8}$ & 0.33207594 \\
			50    & $7.43\times10^{-8}$ & 0.33207603 \\
			60    & $7.28\times10^{-8}$ & 0.33207608 \\
			70    & $7.14\times10^{-8}$ & 0.33207619 \\
			80    & $7.01\times10^{-8}$ & 0.33207621 \\
			90    & $6.89\times10^{-8}$ & 0.33207623 \\
			Final & $6.789\times10^{-8}$ & 0.3320762918 \\
			\bottomrule
		\end{tabular}
	\end{table}
	
	The weighted objective decreases steadily during L-BFGS, while the
	wall-shear prediction remains in a narrow range near $0.332076$.
	Importantly, the reported Adam values at iterations $13000$ and
	$14000$ are closer to the high-accuracy benchmark than the final
	L-BFGS prediction, even though the weighted objective is reduced
	further during L-BFGS.
	
	This demonstrates that minimization of the complete PINN objective
	does not necessarily imply monotonic improvement in a particular
	physical quantity of interest. L-BFGS minimizes the global
	physics-informed objective rather than the error in $f''(0)$ alone.
	
	\subsection{Predicted Blasius Profiles}
	
	The representative adaptive PINN solution evaluated at integer values
	of $\eta$ is given in Table~\ref{tab:profile}.
	
	\begin{table}[H]
		\centering
		\caption{Representative adaptive PINN solution for the Blasius equation.}
		\label{tab:profile}
		\begin{tabular}{cccc}
			\toprule
			$\eta$ & $f(\eta)$ & $f'(\eta)$ & $f''(\eta)$ \\
			\midrule
			0 & 0.000002 & 0.000001 & 0.332076 \\
			1 & 0.165570 & 0.329764 & 0.322968 \\
			2 & 0.649993 & 0.629726 & 0.266733 \\
			3 & 1.396729 & 0.845988 & 0.161346 \\
			4 & 2.305609 & 0.955464 & 0.064253 \\
			5 & 3.283088 & 0.991496 & 0.015899 \\
			6 & 4.279387 & 0.998927 & 0.002419 \\
			7 & 5.278966 & 0.999885 & 0.000217 \\
			8 & 6.278916 & 1.000000 & 0.000055 \\
			\bottomrule
		\end{tabular}
	\end{table}
	
	The normalized velocity increases from approximately zero at the wall
	to the imposed far-field value. At $\eta=5$,
	
	\begin{equation}
		f'(5)=0.991496,
	\end{equation}
	
	so the velocity has already reached more than $99\%$ of the
	free-stream value. The shear decreases from approximately $0.332076$
	at the wall to
	
	\begin{equation}
		f''(8)\approx5.5\times10^{-5},
	\end{equation}
	
	showing the expected reduction of the velocity gradient away from the
	plate.
	
\begin{figure}[H]
		\centering
		\includegraphics[width=0.95\textwidth]{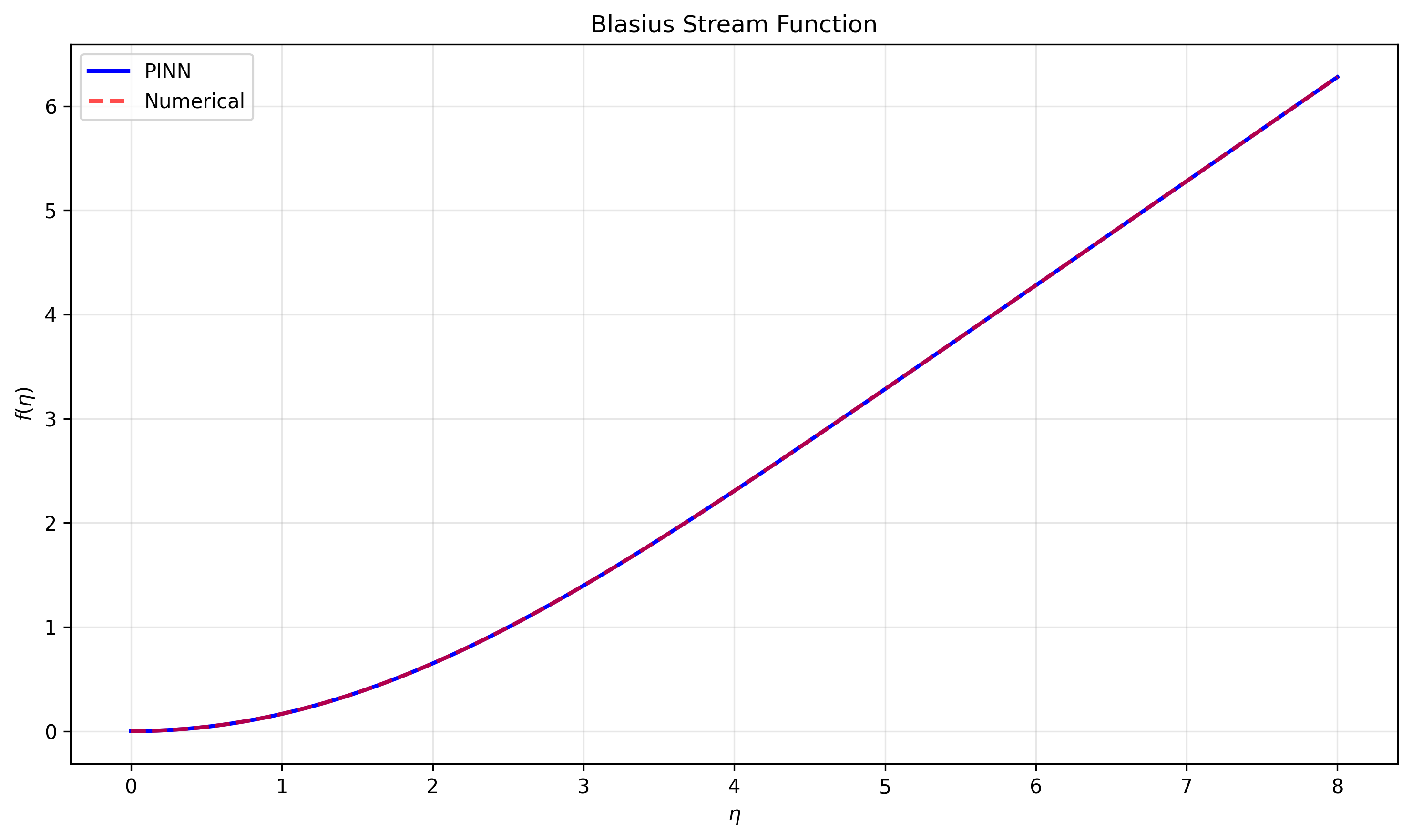}
		\caption{
			Comparison of the Blasius stream function $f(\eta)$ predicted by
			the adaptive PINN and the independent numerical BVP solution.
		}
		\label{fig:f}
	\end{figure}

	Figure~\ref{fig:f} compares the stream-function prediction with the
	independent numerical BVP solution. The two curves are visually almost
	indistinguishable throughout the computational interval.
	
\begin{figure}[H]
		\centering
		\includegraphics[width=0.95\textwidth]{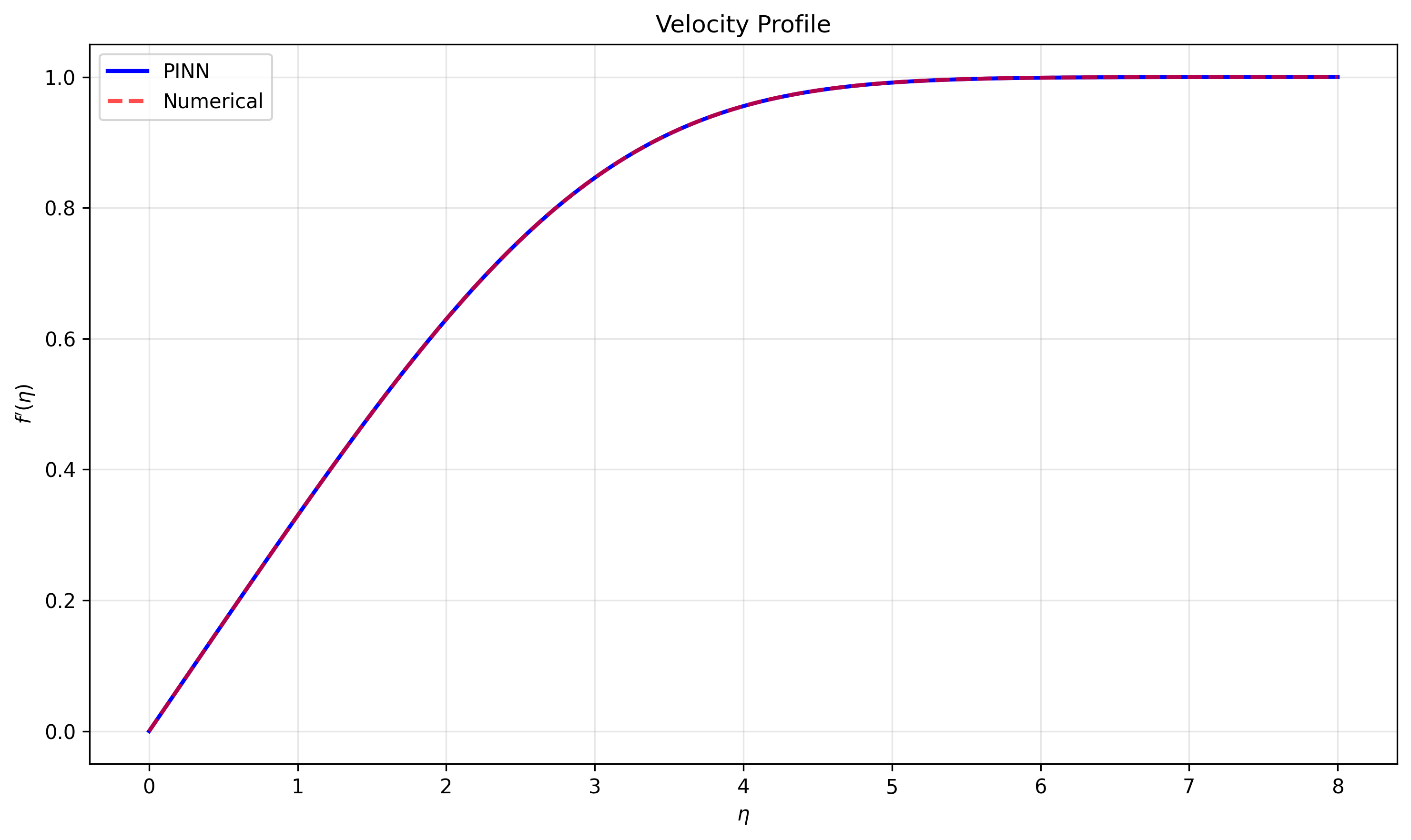}
		\caption{
			Comparison of the nondimensional velocity profile $f'(\eta)$
			predicted by the adaptive PINN and the numerical BVP solution.
		}
		\label{fig:fp}
	\end{figure}

	The nondimensional velocity profiles are shown in
	Fig.~\ref{fig:fp}. The adaptive PINN reproduces the monotonic increase
	of $f'(\eta)$ from the no-slip wall condition toward the free-stream
	value.
	
\begin{figure}[H]
		\centering
		\includegraphics[width=0.95\textwidth]{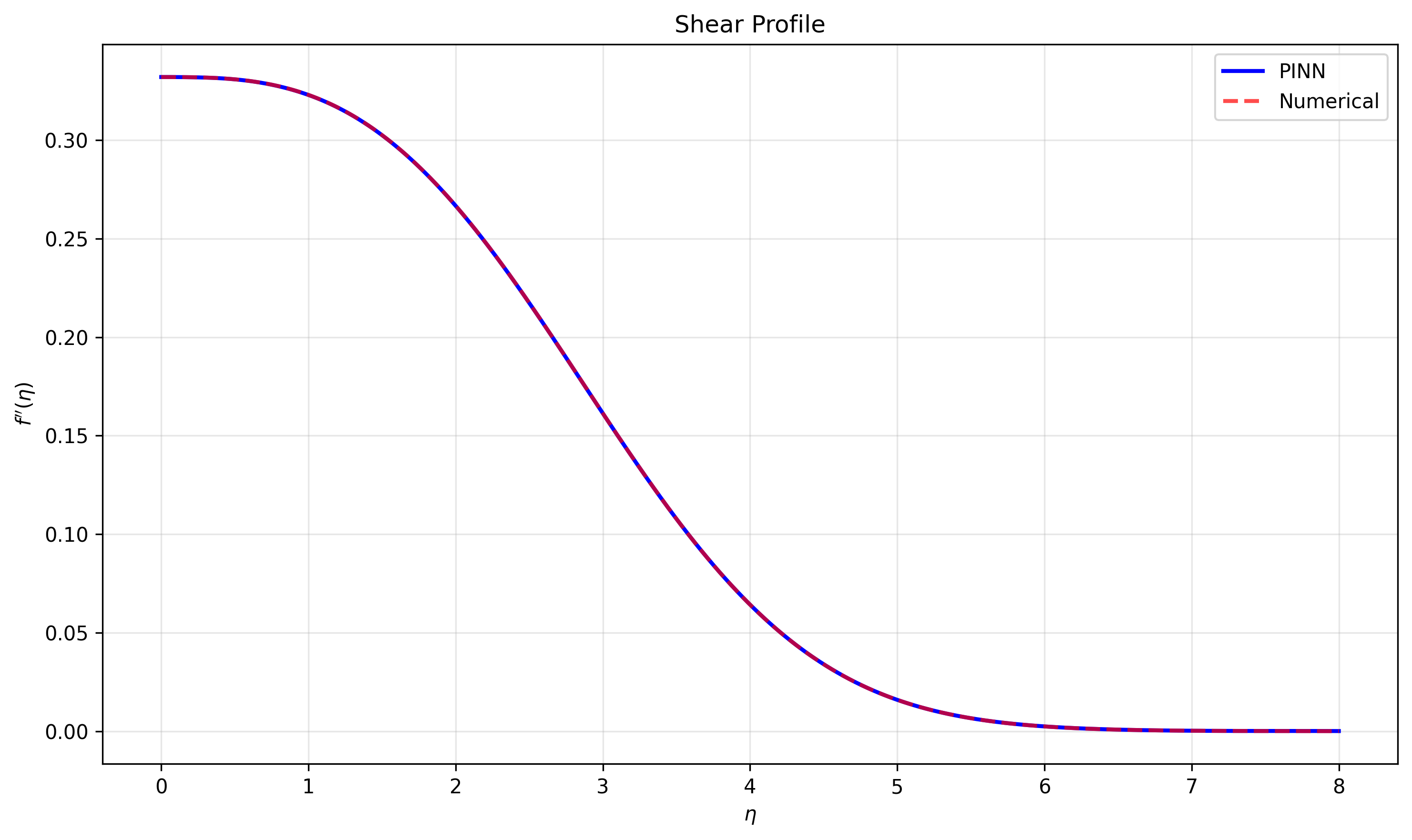}
		\caption{
			Comparison of the shear profile $f''(\eta)$ predicted by the
			adaptive PINN and the numerical BVP solution.
		}
		\label{fig:fpp}
	\end{figure}

	Figure~\ref{fig:fpp} shows the shear profile. Agreement in
	$f''(\eta)$ provides a particularly useful validation measure because
	derivative quantities can expose discrepancies that are less visible
	in the stream-function profile itself.
	
	\subsection{Profile Error and Residual Validation}
	
\begin{figure}[H]
		\centering
		\includegraphics[width=0.95\textwidth]{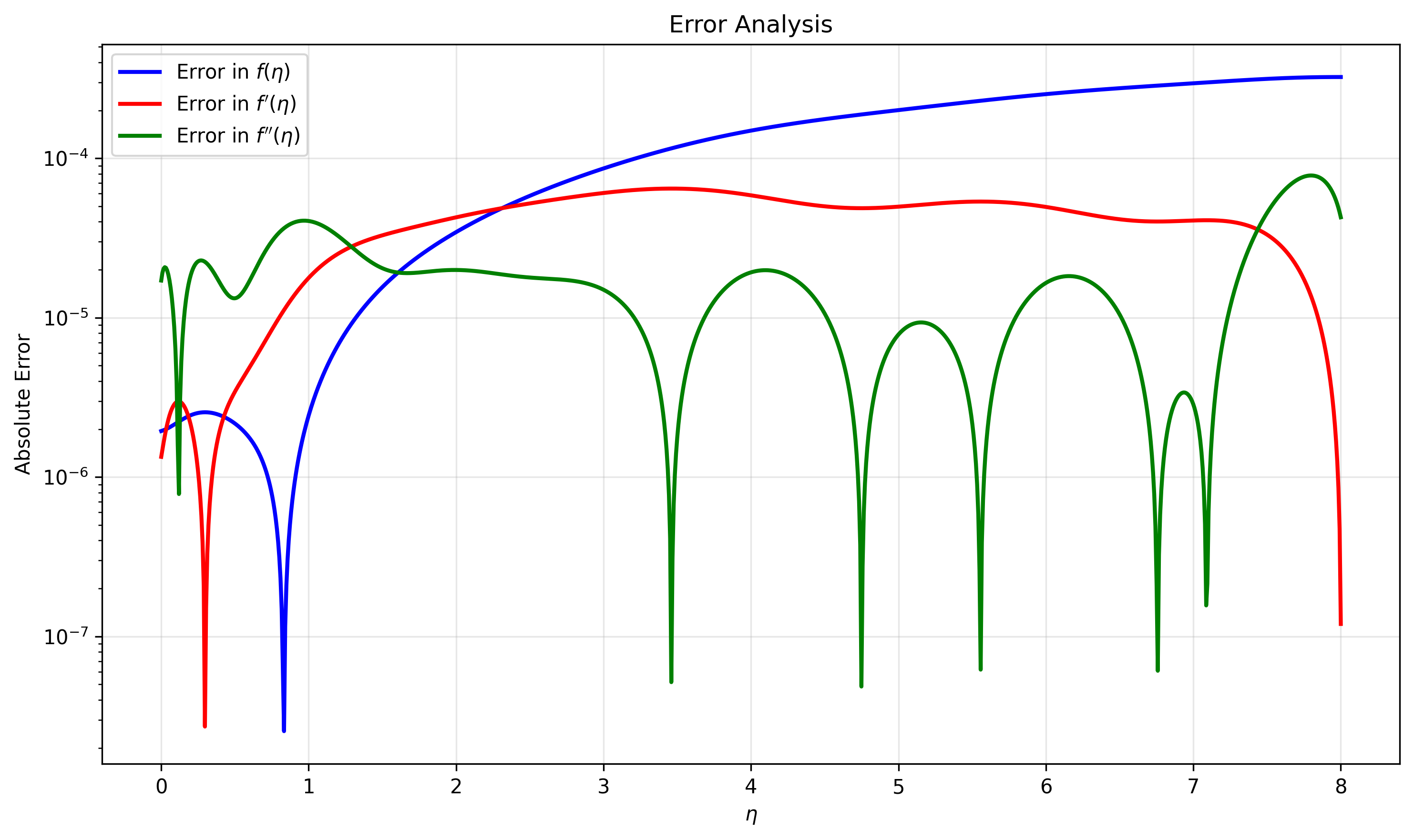}
		\caption{
			Pointwise absolute errors in $f(\eta)$, $f'(\eta)$, and
			$f''(\eta)$ relative to the independent numerical BVP solution.
		}
		\label{fig:error}
	\end{figure}

	Although the direct profiles overlap closely, the pointwise absolute
	error curves in Fig.~\ref{fig:error} reveal the remaining local
	differences between the PINN and the independent BVP solution.
	
	The error in $f(\eta)$ generally increases toward the outer part of
	the computational interval but remains small. The velocity error is
	also small throughout the domain, while the error in $f''(\eta)$
	displays a more oscillatory spatial structure. Sharp local minima in
	the absolute-error curves occur where the corresponding PINN and
	numerical predictions become nearly equal.
	
\begin{figure}[H]
		\centering
		\includegraphics[width=0.95\textwidth]{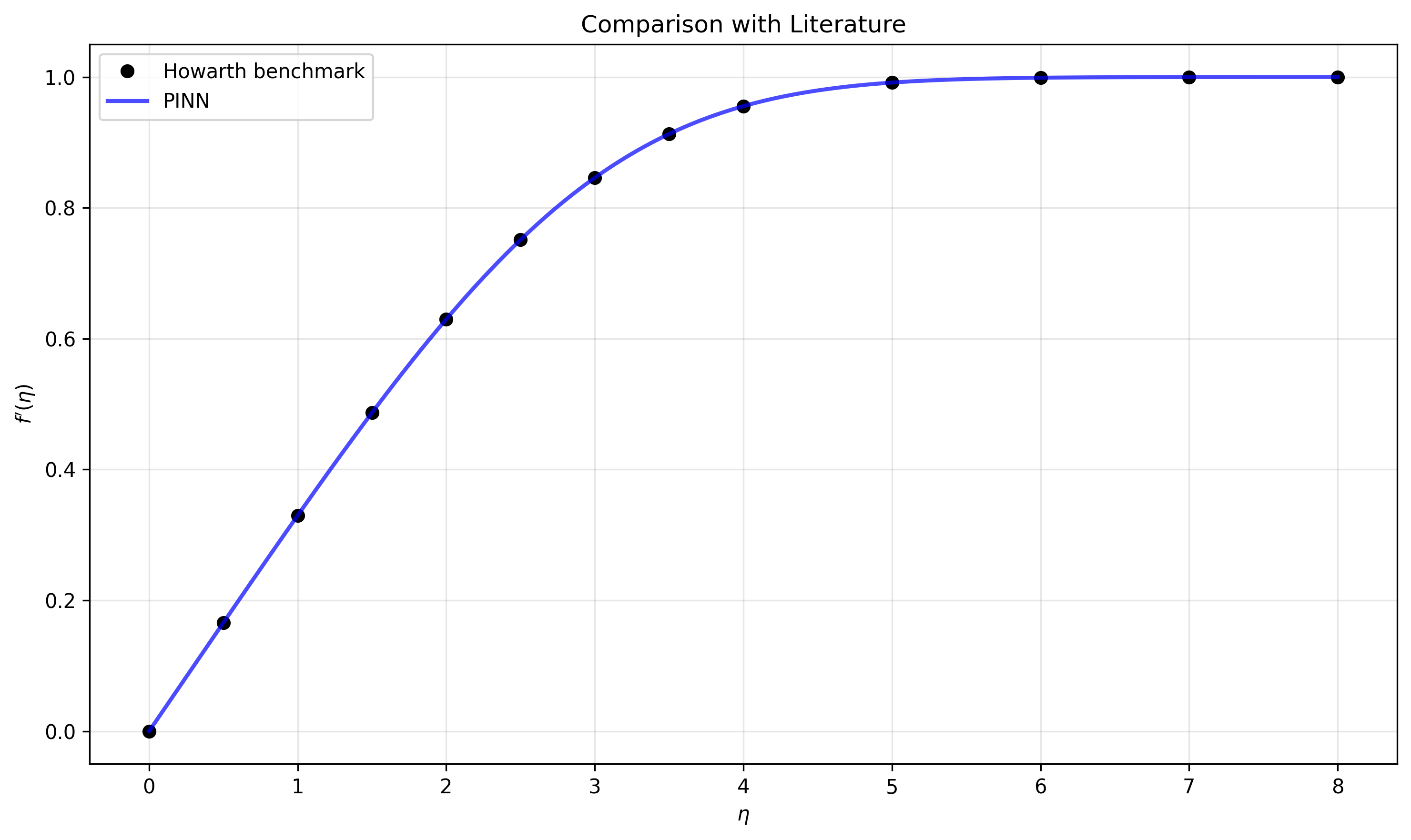}
		\caption{
			Comparison of the adaptive PINN nondimensional velocity profile
			with classical Blasius reference values~\cite{howarth1938}.
		}
		\label{fig:literature}
	\end{figure}

	Figure~\ref{fig:literature} provides an additional comparison between
	the adaptive PINN velocity curve and classical rounded Blasius
	reference values~\cite{howarth1938}. The PINN follows the reference
	profile closely throughout the boundary layer.
	
	A separate physics-based diagnostic is provided by the pointwise
	governing-equation residual,
	
	\begin{equation}
		\mathcal{R}(\eta)
		=
		\left|
		f'''_{\theta}(\eta)
		+
		\frac{1}{2}
		f_{\theta}(\eta)
		f''_{\theta}(\eta)
		\right|.
		\label{eq:absolute_residual}
	\end{equation}
	
	For the reported trained network, the maximum residual evaluated over
	the $500$ plotting points is approximately
	
	\begin{equation}
		\mathcal{R}_{\max}
		\approx
		3.24\times10^{-4}.
	\end{equation}
	
	The residual is nonuniform across the domain and contains localized
	peaks. This behavior provides additional motivation for the
	residual-directed collocation strategy, which allocates new training
	points to locations where the current ODE residual is comparatively
	large.
	
	\begin{figure}[H]
		\centering
		\includegraphics[width=0.95\textwidth]{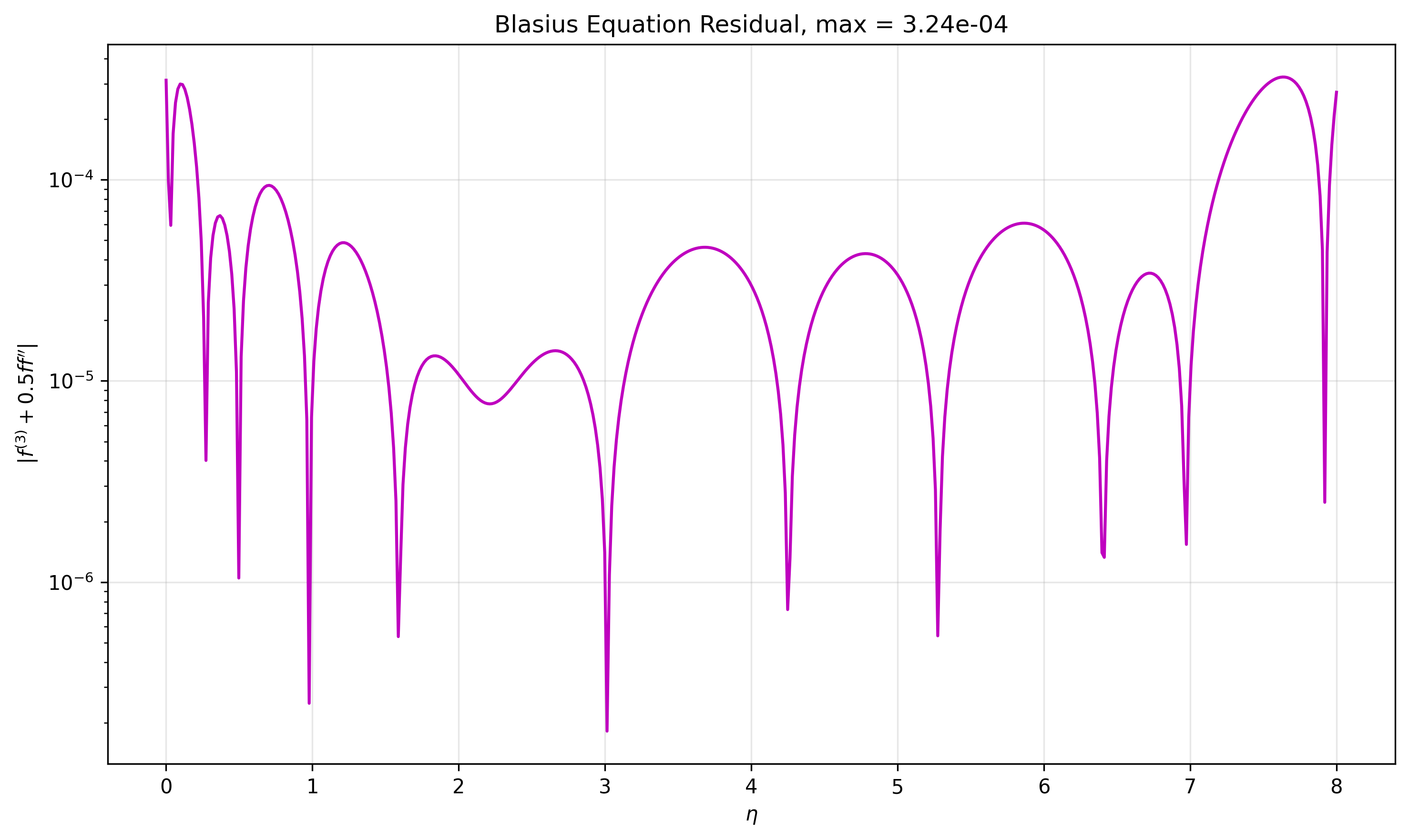}
		\caption{
			Pointwise absolute Blasius-equation residual over the
			computational domain. The maximum residual over the $500$
			evaluation points is approximately $3.24\times10^{-4}$.
		}
		\label{fig:residual}
	\end{figure}
	
	Taken together, the wall-shear benchmark, complete profile comparison,
	pointwise errors, boundary-condition losses, and governing-equation
	residual provide complementary measures of the quality of the trained
	solution.
	
	\section{Architecture Study}
	\label{sec:architecture}
	
	A separate full-training study was performed for four network
	architectures:
	
	\begin{align}
		f_1 &: [1,100,100,1], \\
		f_2 &: [1,100,100,100,1], \\
		f_3 &: [1,90,90,90,90,1], \\
		f_4 &: [1,90,90,90,90,90,1].
	\end{align}
	
	Each model underwent $15000$ Adam iterations, residual-based adaptive
	collocation, restoration of its own best Adam state, and subsequent
	L-BFGS refinement. Model caching was disabled and the full training
	configuration was used for all four models.
	
	The resulting values are summarized in
	Table~\ref{tab:architecture_variation}.
	
	\begin{table}[H]
		\centering
		\caption{
			Full-training results for the four adaptive PINN architectures.
		}
		\label{tab:architecture_variation}
		\resizebox{\textwidth}{!}{
			\begin{tabular}{cccccc}
				\toprule
				\textbf{Model} &
				\textbf{Architecture} &
				$\bm{f''(0)}$ &
				\textbf{Absolute Error} &
				\textbf{Raw ODE Loss} &
				\textbf{Weighted Loss} \\
				\midrule
				
				$f_1$ &
				$[1,100,100,1]$ &
				0.3320557068 &
				$1.629\times10^{-6}$ &
				$4.072\times10^{-8}$ &
				$3.089\times10^{-7}$ \\
				
				$f_2$ &
				$[1,100,100,100,1]$ &
				0.3321201301 &
				$6.279\times10^{-5}$ &
				$1.209\times10^{-8}$ &
				$1.209\times10^{-7}$ \\
				
				$f_3$ &
				$[1,90,90,90,90,1]$ &
				0.3320457565 &
				$1.158\times10^{-5}$ &
				$2.017\times10^{-8}$ &
				$2.017\times10^{-7}$ \\
				
				$f_4$ &
				$[1,90,90,90,90,90,1]$ &
				0.3318284466 &
				$2.289\times10^{-4}$ &
				$2.424\times10^{-7}$ &
				$2.424\times10^{-8}$ \\
				
				\bottomrule
			\end{tabular}
		}
	\end{table}
	
	Among the four tested architectures, $f_1$ produces the most accurate
	wall-shear prediction,
	
	\begin{equation}
		f''(0)=0.3320557068,
	\end{equation}
	
	with an absolute error of
	
	\begin{equation}
		e_{f_1}
		=
		1.629\times10^{-6}.
		\label{eq:f1error}
	\end{equation}
	
	The four-hidden-layer model $f_3$ also gives a relatively small
	wall-shear error,
	
	\begin{equation}
		1.158\times10^{-5}.
	\end{equation}
	
	The corresponding errors for $f_2$ and $f_4$ are
	
	\begin{equation}
		6.279\times10^{-5}
		\qquad\text{and}\qquad
		2.289\times10^{-4},
	\end{equation}
	
	respectively.
	
	In contrast to the wall-shear ranking, the deepest architecture
	$f_4$ has the smallest reported weighted objective,
	
	\begin{equation}
		\mathcal{L}_{f_4}
		=
		2.424\times10^{-8}.
	\end{equation}
	
	It should therefore be described as the model with the smallest
	weighted training objective, rather than as the most physically
	accurate architecture.
	
	The raw ODE losses help explain why weighted objectives must be
	interpreted cautiously. For $f_4$, the raw ODE loss is
	
	\begin{equation}
		2.424\times10^{-7},
	\end{equation}
	
	which is one order of magnitude larger than the reported weighted
	objective. This result is consistent with a small adaptive coefficient
	multiplying the ODE contribution at the end of training. Since each
	architecture evolves its own adaptive coefficients during Adam and
	those coefficients are subsequently held fixed during L-BFGS,
	weighted losses from different architectures do not necessarily
	represent identically scaled objective functions.
	
	Consequently, architecture ranking based solely on weighted loss can
	differ substantially from ranking based on a physical quantity such as
	$f''(0)$. The wall-shear error, raw residual losses, boundary errors,
	and complete solution profiles therefore provide more physically
	meaningful criteria for model assessment.
	
	The representative baseline calculation and the architecture-study
	$f_1$ calculation use the same nominal network configuration but give
	
	\begin{equation}
		f''(0)=0.3320762918
	\end{equation}
	
	and
	
	\begin{equation}
		f''(0)=0.3320557068,
	\end{equation}
	
	respectively. These are separately trained realizations. The random
	seeds are initialized once at the beginning of the program but are not
	reset before each architecture is constructed, so the global
	random-number-generator state advances between model initializations.
	The two values should therefore not be expected to coincide exactly.
	
	\section{Comparison with Previous PINN Results}
	\label{sec:comparison}
	
	Krishna et al.~\cite{krishna2023} provide a direct PINN reference for
	the same Blasius boundary-value problem on the truncated interval
	$0\leq\eta\leq8$. For a network with two hidden layers and $100$
	neurons in each layer, they reported
	
	\begin{equation}
		f''(0)_{\mathrm{Krishna}}
		=
		0.33165
	\end{equation}
	
	with a reported loss of
	
	\begin{equation}
		1.67\times10^{-6}.
	\end{equation}
	
	Using the high-accuracy reference adopted in this study,
	
	\begin{align}
		e_{\mathrm{Krishna}}
		&=
		\left|
		0.33165
		-
		0.332057336215
		\right|
		\\
		&=
		4.07336\times10^{-4}.
	\end{align}
	
	For the representative adaptive PINN,
	
	\begin{equation}
		e_{\mathrm{baseline}}
		=
		1.896\times10^{-5}.
	\end{equation}
	
	The ratio of these wall-shear errors is approximately
	
	\begin{equation}
		\frac{
			e_{\mathrm{Krishna}}
		}{
			e_{\mathrm{baseline}}
		}
		\approx
		21.5.
	\end{equation}
	
	Thus, the representative adaptive PINN gives a wall-shear error
	approximately $21.5$ times smaller than that associated with the
	reported value of Krishna et al.~\cite{krishna2023}.
	
	The separately trained architecture-study $f_1$ model gives an even
	smaller wall-shear error,
	
	\begin{equation}
		e_{f_1}
		=
		1.629\times10^{-6}.
	\end{equation}
	
	This value is approximately $250$ times smaller than the error
	associated with the reported Krishna et al. value. This latter
	comparison should be interpreted as a comparison between resulting
	physical predictions rather than as a controlled architecture
	experiment, because the training methodologies differ.
	
	Table~\ref{tab:comparison_reference} summarizes the main differences
	between the previous Blasius PINN formulation and the representative
	adaptive calculation.
	
	\begin{table}[H]
		\centering
		\caption{
			Comparison with the Blasius PINN formulation of Krishna
			et al.~\cite{krishna2023}.
		}
		\label{tab:comparison_reference}
		\resizebox{\textwidth}{!}{
			\begin{tabular}{p{4.3cm}cc}
				\toprule
				\textbf{Feature} &
				\textbf{Krishna et al.~\cite{krishna2023}} &
				\textbf{Present representative PINN} \\
				\midrule
				
				Computational domain &
				$[0,8]$ &
				$[0,8]$ \\
				
				Representative network &
				2 hidden layers, 100 neurons each &
				$[1,100,100,1]$ \\
				
				Base collocation &
				100 equidistant points &
				499 unique nonuniform points \\
				
				Near-wall concentration &
				Uniform distribution &
				60\% of nominal base points in $[0,3]$ \\
				
				Residual-based refinement &
				Not reported &
				Yes \\
				
				Loss weighting &
				Non-adaptive combined objective &
				Gradient-norm adaptive weighting \\
				
				Automatic differentiation &
				Yes &
				Yes \\
				
				Optimization &
				Adam and L-BFGS &
				Adam and L-BFGS \\
				
				Reported $f''(0)$ &
				0.33165 &
				0.3320762918 \\
				
				Wall-shear error &
				$4.07336\times10^{-4}$ &
				$1.896\times10^{-5}$ \\
				
				Reported final/weighted loss &
				$1.67\times10^{-6}$ &
				$6.789\times10^{-8}$ \\
				
				Profile validation &
				Numerical comparison &
				Independent \texttt{solve\_bvp} comparison \\
				
				Negative-$\eta$ analysis &
				Included &
				Not considered \\
				
				\bottomrule
			\end{tabular}
		}
	\end{table}
	
	The reported loss values in Table~\ref{tab:comparison_reference}
	should not be interpreted as directly comparable accuracy measures.
	The two studies use different collocation densities, sampling
	strategies, loss formulations, weighting procedures, and optimization
	settings. The wall-shear error relative to a common high-accuracy
	reference is therefore a more meaningful scalar comparison.
	
	The scopes of the two investigations are also different. Krishna
	et al.~\cite{krishna2023} extended their PINN solution toward the
	negative-$\eta$ singularity. The present study instead concentrates on
	adaptive loss balancing, residual-directed collocation, staged
	optimization, detailed validation over the standard positive domain,
	and sensitivity to network architecture.
	
	\section{Discussion}
	\label{sec:discussion}
	
	The numerical experiments reveal several important characteristics of
	the adaptive PINN formulation.
	
	First, the adaptive loss coefficients change substantially throughout
	Adam optimization. Figure~\ref{fig:weights} shows that the individual
	physical constraints generate different gradient scales during
	training. Fixed equal coefficients would therefore not necessarily
	produce balanced contributions to the parameter updates.
	
	Second, residual-based collocation enables the training-point
	distribution to respond to the evolving approximation. The nonuniform
	base grid provides stable coverage with additional near-wall
	resolution, while adaptive points are introduced at locations where
	the current differential-equation residual is comparatively large.
	The nonuniform residual distribution in Fig.~\ref{fig:residual}
	supports this strategy.
	
	Third, the training history shows that convergence of an adaptive
	weighted objective need not be monotonic. Periodic reweighting changes
	the relative scaling of the individual terms, while residual-based
	refinement modifies the set of coordinates at which the ODE loss is
	evaluated. Short-term increases in the weighted objective can therefore
	occur within an overall decreasing training trend.
	
	Fourth, Adam and L-BFGS play different numerical roles. Adam generates
	the dominant reduction of the objective and identifies the main
	physical solution. L-BFGS then reduces the weighted objective further
	under the final collocation configuration while producing only small
	changes in $f''(0)$.
	
	A particularly important result is the absence of a one-to-one
	relationship between weighted training loss and wall-shear accuracy.
	Within the representative training trajectory, the late Adam
	predictions at iterations $13000$ and $14000$ are closer to the
	high-accuracy wall-shear reference than the final L-BFGS prediction,
	even though the latter has a lower global objective under its final
	training configuration.
	
	The architecture study provides a second manifestation of the same
	phenomenon. Among the four tested architectures, $f_4$ achieves the
	smallest weighted objective but the largest wall-shear error, whereas
	$f_1$ gives the smallest wall-shear error despite having a larger
	weighted loss. This difference arises partly because $f''(0)$ is a
	localized physical quantity, while the PINN objective measures
	combined behavior over the full computational domain. Moreover,
	different architecture runs finish with independently evolved adaptive
	coefficients, so their weighted objectives are not necessarily
	identically scaled.
	
	These observations support validation using several complementary
	criteria:
	
	\begin{enumerate}[label=(\roman*)]
		\item the physical wall-shear quantity $f''(0)$;
		\item the raw governing-equation residual;
		\item wall and far-field boundary errors;
		\item complete profile agreement with an independent numerical
		solution;
		\item the pointwise residual distribution; and
		\item optimization convergence.
	\end{enumerate}
	
	The comparison with Krishna et al.~\cite{krishna2023} shows a
	substantial reduction in wall-shear error for the representative
	adaptive calculation. However, this improvement should not be
	attributed to any single adaptive mechanism because the two
	calculations differ simultaneously in collocation density,
	collocation placement, loss weighting, numerical precision,
	initialization, and optimization details. A controlled ablation study
	would be required to quantify the separate contributions of
	nonuniform collocation, residual-based refinement, adaptive weighting,
	and staged optimization.
	
	\section{Limitations}
	\label{sec:limitations}
	
	Several limitations of the present study should be recognized. First,
	the investigation is restricted to the standard positive Blasius
	domain truncated at $\eta=8$. The negative-axis singularity examined
	in previous studies~\cite{boyd1999,krishna2023} is not investigated.
	
	Second, a controlled ablation analysis is not performed. Consequently,
	the observed improvement cannot be assigned independently to adaptive
	loss weighting, nonuniform sampling, residual-based refinement, or
	optimizer staging.
	
	Third, each architecture in Section~\ref{sec:architecture} is evaluated
	using one sequential training realization. Since the random-number
	generator is initialized once and not reset before each architecture,
	network architecture and initialization effects are not fully
	separated. Multiple runs with separately controlled seeds would be
	required for a statistical comparison of architecture performance.
	
	Fourth, the final weighted objectives of different architectures are
	influenced by independently evolved adaptive coefficients and should
	therefore not be interpreted as identically normalized quantities.
	
	Finally, the present work emphasizes solution accuracy rather than
	computational cost. No wall-clock comparison with shooting,
	finite-difference, or other classical numerical methods is included.
	Such benchmarking would be required before making conclusions about
	relative computational efficiency.
	
	\section{Conclusion}
	\label{sec:conclusion}
	
	An adaptive physics-informed neural network was developed for the
	Blasius boundary-layer equation using gradient-norm-based loss
	weighting, nonuniform and residual-based collocation, and sequential
	Adam--L-BFGS optimization. For the representative architecture
	$[1,100,100,1]$, the model predicts
	$f''(0)=0.3320762918$, compared with the high-accuracy reference value
	$0.332057336215$, giving an absolute error of
	$1.896\times10^{-5}$. The final weighted loss is
	$6.789\times10^{-8}$, and the predicted stream-function, velocity, and
	shear profiles agree closely with the independent numerical BVP
	solution.
	
	The architecture study further shows that the smallest weighted loss
	does not necessarily correspond to the most accurate physical result.
	Among the four tested architectures, the two-hidden-layer model
	$f_1=[1,100,100,1]$ achieves the smallest wall-shear error,
	$1.629\times10^{-6}$, whereas the deepest model $f_4$ obtains the
	smallest weighted objective but a larger wall-shear error of
	$2.289\times10^{-4}$. PINN performance should therefore be evaluated
	using physically meaningful quantities, raw residuals, and
	boundary-condition errors rather than weighted optimization loss
	alone.
	
	Compared with the previously reported PINN value
	$f''(0)=0.33165$ of Krishna et al.~\cite{krishna2023}, the
	representative adaptive formulation provides substantially improved
	wall-shear accuracy. Overall, the results indicate that adaptive loss
	balancing, residual-directed collocation, and staged optimization
	provide an effective framework for accurately solving the Blasius
	problem.
	

\end{document}